\documentclass[lettersize,journal]{IEEEtran}

\usepackage{amsmath,amsfonts}
\usepackage{graphicx}
\usepackage{enumitem} 
\usepackage{amsmath}   
\usepackage{amsfonts} 
\usepackage{booktabs}  
\usepackage{listings}  
\usepackage{xcolor}    
\usepackage{color}   
\usepackage{colortbl}
\usepackage{subcaption}
\usepackage[utf8]{inputenc}
\usepackage{multirow}
\usepackage{hyperref}

\usepackage{algorithm}%
\usepackage{algorithmicx}%
\usepackage{algpseudocode}%

\definecolor{codegreen}{rgb}{0,0.4,0}
\definecolor{codegray}{rgb}{0.5,0.5,0.5}
\definecolor{codepurple}{rgb}{0.58,0,0.82}
\definecolor{backcolour}{rgb}{1,1,1}
\definecolor{codeblue}{rgb}{0.96,0.98,1}
\definecolor{highlight}{rgb}{1,0.96,0.98}

\lstdefinestyle{mystyle}{
  backgroundcolor=\color{codeblue}, commentstyle=\color{codegreen},
  keywordstyle=\color{red},
  numberstyle=\tiny\color{codegray},
  stringstyle=\color{codepurple},
  basicstyle=\ttfamily\footnotesize,
  breakatwhitespace=false,         
  breaklines=true,                 
  captionpos=b,                    
  keepspaces=true,                 
  numbers=left,                    
  numbersep=5pt,                  
  showspaces=false,                
  showstringspaces=false,
  showtabs=false,                  
  tabsize=2,
}
\begin{document}

\title{{Interpreting and Improving Attention From the Perspective of Large Kernel Convolution}}

\author{
Chenghao Li, Chaoning Zhang,~\IEEEmembership{Senior,~IEEE,}  Boheng Zeng, Yi Lu,\\ Pengbo Shi, Qingzi Chen, Jirui Liu, Lingyun Zhu,\\ Yang Yang,~\IEEEmembership{Senior,~IEEE}, Heng Tao Shen,~\IEEEmembership{Fellow,~IEEE}
\thanks{Chenghao Li and Chaoning Zhang are with the School of Computing, Kyung Hee University (e-mail: lch17692405449@gmail.com; chaoningzhang1990@gmail.com;). Boheng Zeng, Yang Yang, and Heng Tao Shen are with the School of Computer Science and Engineering, University of Electronic Science and Technology, China (e-mail: zengboheng@std.uestc.edu.cn; yang.yang@uestc.edu.cn; shenhengtao@hotmail.com); Yi Lu is with the Capital Normal University (e-mail: 2230501004@cnu.edu.cn;). Pengbo Shi, Qingzi Chen, Jirui Liu, and Lingyun Zhu are with the Chongqing University of Technology, China (e-mail: shipengbo@cqut.edu.cn; chenqingzi@cqut.edu.cn; liujirui@cqut.edu.cn; zly69cv@163.com;)}}

\markboth{Journal of \LaTeX\ Class Files,~Vol.~14, No.~8, August~2015}%
{Shell \MakeLowercase{\textit{et al.}}: FlashFormer: Efficient ViT Arch without Hybridizing CNNs}

\maketitle

\begin{abstract}

Attention mechanisms have significantly advanced visual models by capturing global context effectively. However, their reliance on large-scale datasets and substantial computational resources poses challenges in data-scarce and resource-constrained scenarios. Moreover, traditional self-attention mechanisms lack inherent spatial inductive biases, making them suboptimal for modeling local features critical to tasks involving smaller datasets.
In this work, we introduce Large Kernel Convolutional Attention (LKCA), a novel formulation that reinterprets attention operations as a single large-kernel convolution. This design unifies the strengths of convolutional architectures—locality and translation invariance—with the global context modeling capabilities of self-attention. By embedding these properties into a computationally efficient framework, LKCA addresses key limitations of traditional attention mechanisms.
The proposed LKCA achieves competitive performance across various visual tasks, particularly in data-constrained settings. Experimental results on CIFAR-10, CIFAR-100, SVHN, and Tiny-ImageNet demonstrate its ability to excel in image classification, outperforming conventional attention mechanisms and vision transformers in compact model settings. These findings highlight the effectiveness of LKCA in bridging local and global feature modeling, offering a practical and robust solution for real-world applications with limited data and resources.

\end{abstract}

\begin{IEEEkeywords}
ConvNet, Vision Transformer, Large Kernel CNN, Attention Mechanism
\end{IEEEkeywords}

\section{Introduction}

\IEEEPARstart{V}{ision} Transformers (ViTs) have emerged as a powerful backbone network in modern visual models, introducing a paradigm shift from traditional convolutional architectures~\cite{dosovitskiy2020image}. By leveraging self-attention mechanisms, ViTs excel at capturing long-range dependencies and global context, achieving state-of-the-art performance in resource-intensive tasks~\cite{yuan2022volo,wang2021pyramid,he2022masked,jiao2023dilateformer,zhu2022multimodal,chen2024hitfusion,deng2021extended,xu2023fine,ma2023transformer}, such as large-scale image classification\cite{yuan2022volo}, high-resolution segmentation~\cite{wang2021pyramid}, and advanced feature learning~\cite{he2022masked}. However, their effectiveness often comes with substantial computational and data requirements, making them less ideal for data-constrained scenarios~\cite{he2022masked,lee2021vision}.

\begin{figure*}
    \centering
    \includegraphics[width=\linewidth]{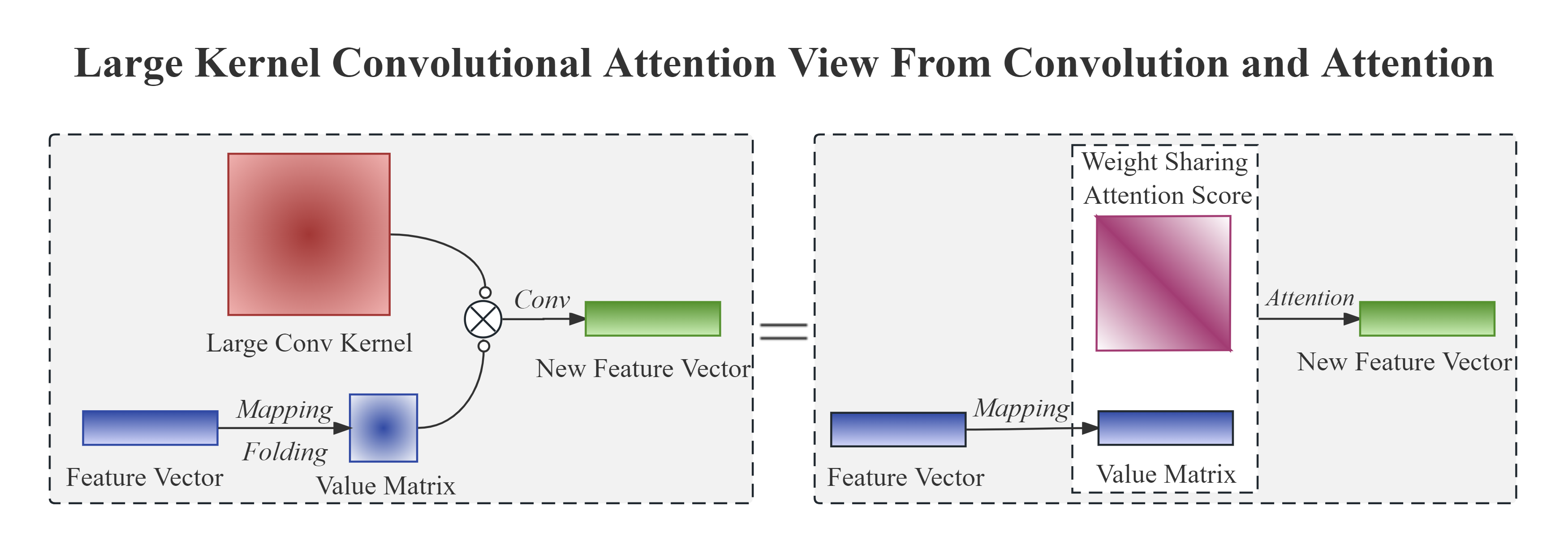}
    \caption{\textbf{Two views to interpret LKCA.} The Large Kernel Convolutional Attention (LKCA) can be understood From the perspective of convolution on the left and attention on the right. The effects of the two approaches are equivalent.}
    \label{fig:enter-label}
\end{figure*}

Conversely, Convolutional Neural Networks (CNNs) have remained a robust choice in environments with limited computational resources or data availability~\cite{he2016deep,krizhevsky2012imagenet}. Due to their strong inductive biases, such as spatial locality and translation invariance, CNNs are inherently efficient in learning from smaller datasets and exhibit robustness to subtle variations in object positions and orientations~\cite{sermanet2013overfeat}. These properties have made CNNs a go-to solution for tasks in low-resource settings, including embedded vision applications and small-scale datasets.
Given their complementary strengths, CNNs can be employed to enhance the performance of ViTs, particularly in data-scarce environments. For instance, CNNs can serve as efficient feature extractors to preprocess input data for ViTs, reducing the computational load on the transformer layers~\cite{hassani2021escaping}. Additionally, hybrid architectures that integrate convolutional modules within ViTs have demonstrated promise in alleviating the limitations of ViTs, such as their lack of inherent spatial inductive bias~\cite{hassani2021escaping}, while retaining their ability to capture global context~\cite{raghu2021vision,jia2021scaling}. This synergy enables the design of models that balance computational efficiency with high-performance capabilities, paving the way for their deployment across a wider range of visual tasks, from resource-constrained devices to large-scale systems.

In this work, we introduce a mechanism called Large Kernel Convolutional Attention (LKCA), designed to effectively combine the strengths of CNNs and ViTs on small datasets. Following the ViT structure, LKCA utilizes key techniques such as Patch Embedding, Positional Encoding, Multi-Head Self-Attention, Feedforward Networks in Transformer Encoder layers, and MLP Head for task output. To effectively incorporate the benefits of CNNs, we employ a simple yet highly efficient operation in LKCA: replacing the conventional self-attention operation in ViT with a single large-kernel convolution operation. This approach not only retains the advantages of translation invariance and spatial inductive bias in CNNs but also leverages the global modeling capability and long-range dependency handling of ViTs. With this large-kernel convolution, LKCA captures both local and global information more effectively, enhancing the model's representational power. From the perspective of attention mechanisms, LKCA can be viewed as an innovative spatial attention method. It efficiently captures spatial features within local regions through the large-kernel convolution operation while also benefiting from the global modeling capabilities of ViT, thus significantly improving the model’s ability to understand and process complex structures in images.
The approach most closely related to our work is the Visual Attention Network (VAN)\cite{guo2023visual}, which replaces the multi-head self-attention mechanism with three types of small kernel convolutions: DW-Conv, DW-DConv, and 1$\times$1-Conv. By contrast, our method is simpler, employing a single large kernel convolution\cite{ding2022scaling} to replace MHSA.
We primarily benchmarked small and medium-sized models, as it was previously believed that ViT outperforms CNNs on large datasets and large models, while CNNs perform better than ViTs on ultra-small and specialized datasets. On the CIFAR-10 dataset, LKCA demonstrate superior improvements over the standard ViT across various experimental configurations.

More importantly, the LKCA design performs exceptionally well on downstream tasks. Firstly, in the image classification task, we conducted experiments on classic datasets, including CIFAR-10, CIFAR-100, SVHN, and Tiny-ImageNet, demonstrating the efficacy of LKCA across these datasets. Additionally, LKCA achieves remarkable results in semantic segmentation tasks, particularly with the UPerNet and FPN algorithms. On the ADE20K dataset, LKCA outperforms other models in terms of both aAcc and mIoU after 10,000 and 20,000 iterations. Comparisons with multiple backbone networks further verify the superior performance of LKCA in both parameter efficiency and accuracy. We attribute the high performance of LKCA to the complementary strengths of CNNs and ViTs. By combining CNN's proficiency in local feature extraction with ViT's ability to model global context, LKCA achieves exceptional performance across various visual tasks.

In this work, our main contributions are as follows:  
\begin{itemize}  
    \item We propose Large Kernel Convolutional Attention (LKCA), a novel mechanism that integrates large-kernel convolution into the ViT architecture to combine the local feature extraction advantages of CNNs with the global modeling capabilities of ViTs.  
    \item We demonstrate that LKCA achieves strong performance across various tasks, including image classification on datasets such as CIFAR-10, CIFAR-100, SVHN, and Tiny-ImageNet, as well as semantic segmentation on ADE20K, outperforming several existing backbone architectures in terms of accuracy and efficiency.  
    \item We address the limitations of both CNNs and ViTs by focusing on small and medium-sized models, offering an effective solution for data-limited and resource-constrained scenarios.  

\end{itemize}


\section{Related Works}

\subsection{Attention Mechanisms in Visual Models}

Attention mechanisms~\cite{nadaraya1964estimating,watson1964smooth,galassi2020attention,cho2015describing,wang2016survey,guo2022attention,yang2020image}, originally inspired by the way humans subconsciously focus on and prioritize important aspects of their surroundings, have become a cornerstone in modern machine learning and deep learning research. These mechanisms mimic the human ability to allocate cognitive resources dynamically, making them highly effective in diverse tasks across Natural Language Processing (NLP), computer vision, and beyond. Visual attention mechanisms~\cite{wang2016survey,guo2022attention}, a critical subset of these techniques, can be conceptualized as a process of dynamically assigning weights to specific features within input images. This weighting mechanism helps models to focus on the most relevant parts of an image, significantly improving the interpretability and performance of various computer vision models.
Visual attention methods are diverse and can be broadly classified into four primary categories: \textit{channel attention}\cite{wang2020eca,zhang2018context,chen2019you,shi2020spsequencenet}, \textit{spatial attention}\cite{yuan2018ocnet,zhang2019self,mnih2014recurrent}, \textit{temporal attention}\cite{zhang2019scan,chen2018video}, and \textit{branch attention}\cite{li2019selective,zhang2022resnest,chen2020dynamic}. These categories often appear in conjunction with other innovative combinations~\cite{chen2017sca,park2018bam,wang2017residual,misra2021rotate,fu2019dual,li2018harmonious,yan2019stat}, reflecting the increasing sophistication of attention-based methodologies.
Non-Local Network~\cite{wang2018non} successfully employs self-attention for the first time to simulate non-local relationships in computer vision. CBAM~\cite{woo2018cbam} infers attention weights sequentially along both spatial and channel dimensions and can seamlessly integrate into any CNN architecture. ECA~\cite{wang2020eca} module introduces a non-reductive local cross-channel interaction strategy, which efficiently enhances the attention module's effectiveness through the effective implementation of 1D convolution. EncNet~\cite{zhang2018context} significantly improves the spatial resolution of pixel-level semantic segmentation labels by introducing a context encoding module. OCNet~\cite{yuan2018ocnet} combines an interleaved sparse self-attention scheme with a traditional multi-scale context approach, effectively modeling dense relationships between pixels and providing richer contextual information. SAGAN~\cite{zhang2019self} employs spatial attention to model long-range dependencies in image generation tasks. SCAN~\cite{zhang2019scan} achieves significantly superior performance in Video Person Re-Identification by refining both intra-sequence and inter-sequence feature representations through the introduction of temporal attention to video sequences. SKNet~\cite{li2019selective} enhances the model's capability to capture objects at different scales by introducing branch attention to merge multiple branches with different kernel sizes.

\subsection{ConvNets With Large Kernel }

Compared to traditional small-kernel CNNs~\cite{he2016deep,huang2017densely}, which have dominated deep learning models due to their efficiency and simplicity, large-kernel CNNs~\cite{szegedy2017inception,szegedy2016rethinking,szegedy2015going,simonyan2014very,peng2017large,hu2019local,trockman2022patches,ding2022scaling,guo2023visual} offer a distinct advantage with their broader effective receptive field and stronger preferences for capturing global shape information. These characteristics make large-kernel models particularly appealing for tasks requiring extensive contextual understanding, such as segmentation and recognition. Despite these advantages, early large-kernel models, such as the Inception family~\cite{szegedy2017inception,szegedy2016rethinking,szegedy2015going}, gradually lost favor following the success of architectures like VGG-Net~\cite{simonyan2014very}, which demonstrated the efficiency of stacking smaller kernels. This shift in focus led to a temporary decline in the adoption of large-kernel designs in mainstream applications. A notable research outcome is the Global Convolutional Network (GCN)~\cite{peng2017large}, which employs very large convolutional kernels of 1$\times$K and K$\times$1 to optimize semantic segmentation tasks. However, there are reports suggesting that large kernels may compromise ImageNet performance. On the other hand, the Local Relation Network (LRNet)~\cite{hu2019local} introduces a spatial aggregation operator (LRLayer) as a replacement for standard convolution, which can be seen as a dynamic convolution. ConvMixer~\cite{trockman2022patches}, on the contrary, uses convolutional kernels as large as 9$\times$9 to replace the "mixer" components of ViT or MLP. Inspired by ViT, RepLKNet~\cite{ding2022scaling} utilizes a large 31$\times$31 convolutional kernel, and its outstanding performance is primarily attributed to the effective receptive fields constructed through large kernels. It has been demonstrated that RepLKNet better exploits shape information compared to traditional CNNs. The Visual Attention Network (VAN)~\cite{guo2023visual} employs three types of convolutions—DW-Conv, DW-DConv, and 1$\times$1 Conv—to expand the receptive field of convolutions and integrate them into the attention mechanism.

\subsection{ViTs with Large ERF}

The strength of ViTs lies in their ability to construct large \textit{Effective Receptive Fields (ERFs)}, enabling them to capture global dependencies effectively. Initially, the Transformer architecture~\cite{vaswani2017attention} was developed for NLP, revolutionizing the way sequential data was modeled through its self-attention mechanism. Its migration to the domain of Computer Vision began with ViT~\cite{dosovitskiy2020image}, which demonstrated the powerful representation capabilities of transformers in handling image data. ViT divides images into patches and treats them as sequences, allowing the transformer architecture to process images in a manner analogous to text in NLP.
Building on this foundation, numerous ViT variants~\cite{han2021transformer,liu2021swin,dong2022cswin,chu2021twins,lin2022cat,huang2021shuffle,fang2022msg,chen2021regionvit,zhou2021deepvit,wang2022kvt} have been proposed to enhance performance on a wide range of visual tasks. These variants have introduced various improvements aimed at addressing the unique challenges posed by image data, particularly the need to model both global and local dependencies effectively. The global modeling capabilities of ViTs make them exceptionally well-suited for capturing long-range relationships between image patches, effectively creating large receptive fields. However, the lack of inherent mechanisms for modeling short-term relationships, as found in CNNs, has been a notable limitation of ViTs. CNNs excel at capturing local dependencies through their spatially constrained convolutional operations, a property that ViTs inherently lack due to their patch-based processing.
To address these challenges, researchers have focused on enhancing ViT's ability to model local information, leading to innovative designs and architectures. 
Transformer in Transformer (TNT)~\cite{han2021transformer} enhances the feature representation capability of local patches through internal attention. Swin Transformer~\cite{liu2021swin,dong2022cswin} addresses local modeling challenges in adapting transformers from language to vision by introducing a hierarchical architecture with shift windowing. 
Twins~\cite{chu2021twins} and CAT~\cite{lin2022cat} alternately perform local and global attention layers layer-wise. Shuffle Transformer~\cite{huang2021shuffle,fang2022msg} efficiently establishes inter-window connections through spatial shuffling with deep convolution and enhances short-term dependency modeling with neighboring window connections. RegionViT~\cite{chen2021regionvit} introduces regions and local tokens with different patch sizes, enhancing local modeling through a region-to-local attention mechanism.


\section{Approach}

\subsection{Preliminary}

\subsubsection{Review of Large Kernel Convolution} 
\begin{figure*}
    \centering
    \includegraphics[width=0.75\linewidth]{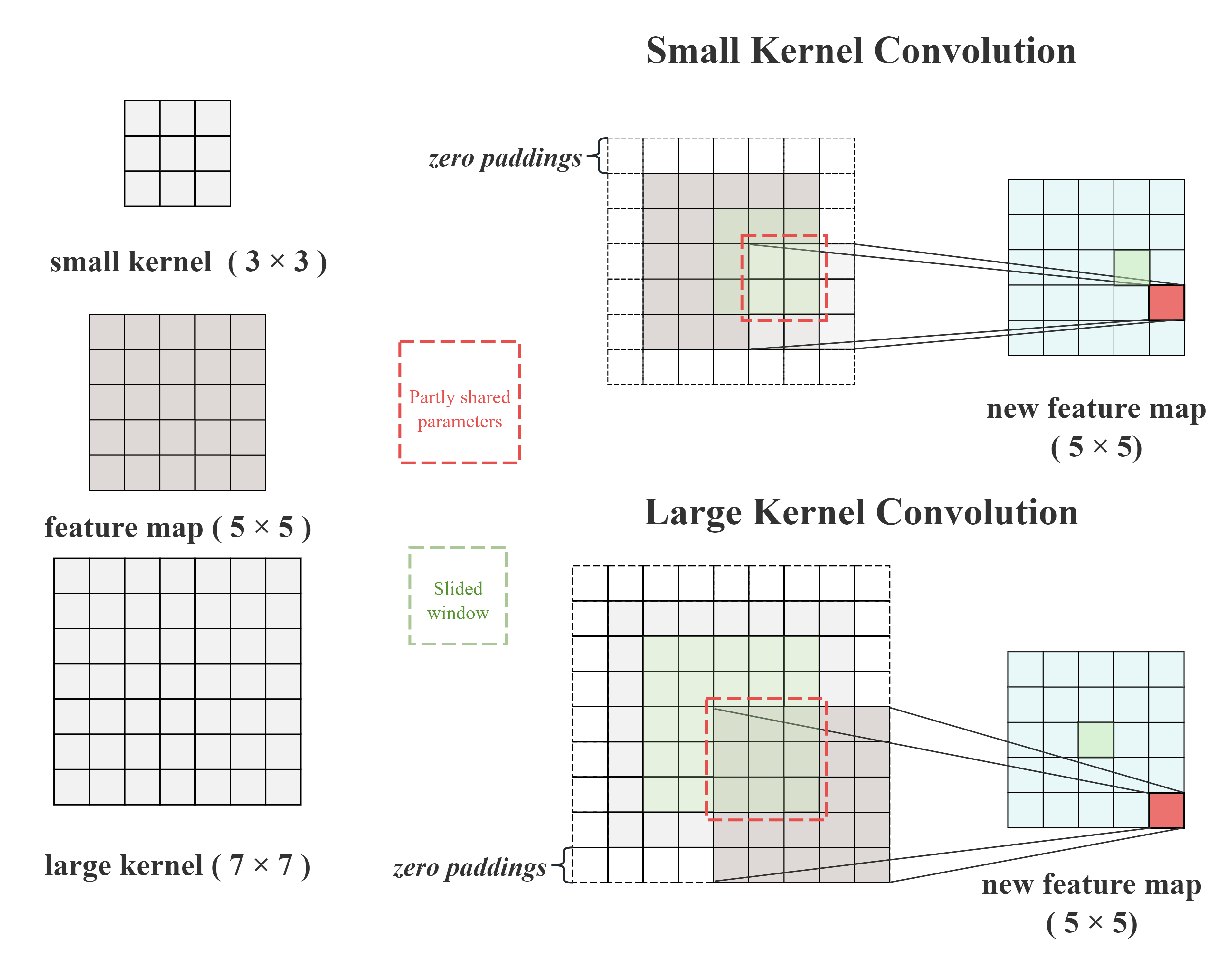}
    \caption{\textbf{Participation differences in kernel convolutions.} Illustration of the difference between small kernel convolution and large kernel convolution by constructing a 5x5 feature map, a 3x3 convolution kernel smaller than the feature map, and a 7x7 convolution kernel larger than the feature map. The distinction between small kernel convolution and large kernel convolution lies in the fact that, in small kernel convolution, all parameters of the kernel are involved in each correlation operation, while only a subset of feature map parameters participates in the computation. In the case of large kernel convolution, during each correlation operation, all parameters of the feature map are involved, and only a subset of the convolution kernel parameters participates in the computation.
}
    \label{fig:largekernel}
\end{figure*}

Ding \textit{et al.}~\cite{ding2022scaling} revisit the paradigm of large kernel design in modern CNNs, drawing inspiration from the ViT. Ding \textit{et al.} assert that employing a few large convolutional kernels instead of multiple smaller ones constitutes a more powerful design paradigm. Building upon this notion, the author introduces a novel CNN architecture named RepLKNet~\cite{ding2022scaling}, featuring kernel sizes as large as 31$\times$31. In comparison to commonly used 3$\times$3 kernels, RepLKNet achieves performance results equivalent to or better than the Swin Transformer~\cite{liu2021swin}, with lower latency. The difference between small kernel convolution and large kernel convolution is shown in Figure~\ref{fig:largekernel}.

\subsubsection{Comparision of LKC and MHSA} Convolution possesses a natural advantage wherein parameters are shared based on geographical location. Specifically, identical parameters of the same convolutional kernel act on every position within the receptive field during the sliding window process. This introduces equivariance and translation invariance to the convolutional operation. However, in Vision Transformer, the attention scores generated by the self-attention mechanism do not explicitly exhibit positional invariance. Attention weights are determined by the feature vectors themselves and the introduced positional encodings. While this enhances the model's fitting capability, it significantly increases the complexity of training. Introducing a mechanism for sharing parameters in the attention mechanism of ViT may bring the expected benefits but without increasing optimization complexity.

\subsection{Large Kernel Convolutional Attention (LKCA)}

\subsubsection{Shared Weight Position Operation} Firstly, we initialize a square matrix of size $N^{\frac{1}{2}} \times 2 - 1$ as learnable parameters, denoted as the \textit{shared parameter weights}, where $N$ represents the number of patches, akin to the initialization of convolutional kernels. Then, we initialize a window of size $N^{\frac{1}{2}}$ and traverse this window as it slides over the \textit{shared parameter weights} matrix from the bottom-right corner to the top-right corner. At each sliding step, the matrix within the window is flattened row-wise into a row vector of size $(1, N)$ and stored. This process is iterated as the window slides, concatenating the row vectors obtained from each step. In total, the window slides $N$ times. Consequently, the concatenated matrix assumes a shape of $(N, N)$, aligning with the shape of the attention weight matrix. This concatenated matrix serves as the new attention score, undergoing matrix multiplication with the linearly mapped Value matrix. This step is referred to as the Shared Weight Matrix Attention. Algorithm~\ref{alg:alg1} presents pseudocode for implementing LKCA from the attention perspective.

\begin{algorithm}
\caption{Forward propagation process of Attention-ViT}
\begin{algorithmic}[1]
\Function{Attention\_ViT\_Forward}{$X$}
    \State \textbf{input:} $X$ -- input tokens, where $X \in \mathbb{R}^{b \times n \times d}$
    \State \textbf{initialize:} $H, W \leftarrow \sqrt{n}$, $weight \leftarrow$ zero tensor of shape $(H*2-1, W*2-1)$
    \State $V \leftarrow \text{Linear}(X)$
    \For{$i = 0$ to $H-1$}
        \For{$j = 0$ to $W-1$}
            \State $start\_row \leftarrow H - 1 - i$
            \State $start\_col \leftarrow W - 1 - j$
            \State $window \leftarrow weight[start\_row:start\_row+H, start\_col:start\_col+W]$
            \State $window\_vector \leftarrow \text{flatten}(window)$
            \State $attn\_score[i*W+j, :] \leftarrow window\_vector$
        \EndFor
    \EndFor
    \State $Out \leftarrow attn\_score @ V$
    \State \Return $Out$
\EndFunction
\end{algorithmic}
\label{alg:alg1}
\end{algorithm}

\subsubsection{From the Perspective of Convolution: Implementing LKCA}

Large Kernel Convolutional Attention (LKCA) can be expressed equivalently as the attention operation with shared parameters described above in terms of code and mathematical implementation. Algorithm~\ref{alg:alg2} showcases pseudocode for implementing LKCA from the perspective of convolution. The convolutional kernel is a learnable weight matrix of size $N^{\frac{1}{2}} \times 2 - 1$, and the input feature map is a matrix $V$ of size $N^{\frac{1}{2}}$ with a channel count equal to the hidden dimension. After zero-padding of size $N^{\frac{1}{2}} - 1$, the convolution operation is applied to the two matrices.

$$\mathbf{LKCA(x)}\mathbf{=}\mathbf{Conv2d}(\mathbf{MLP(x)}, \mathbf{Large Kernel})$$

\begin{algorithm}
\caption{Forward propagation process of Attention-CNN}
\begin{algorithmic}[1]
\Function{Attention\_CNN\_Forward}{$X$}
    \State \textbf{input:} $X$ -- input tokens, where $X \in \mathbb{R}^{b \times n \times d}$
    \State \textbf{initialize:} $H, W \leftarrow \sqrt{n}$, $weight \leftarrow$ zero tensor of shape $(H*2-1, W*2-1)$
    \State $V \leftarrow \text{Linear}(X)$
    \State $V \leftarrow \text{rearrange}(V, 'b (H W) d \rightarrow (b d) H W')$
    \State $Out \leftarrow \text{conv2d}(V, weight, \text{zero\_padding} = (H-1, W-1))$
    \State $Out \leftarrow \text{rearrange}(Out, '(b d) H W \rightarrow b (H W) d')$
    \State \Return $Out$
\EndFunction
\end{algorithmic}
\label{alg:alg2}
\end{algorithm}

\subsubsection{Overall Architecture}

The incorporation of large-kernel convolution attention enhances the model's ability to capture positional information. However, it simultaneously weakens the relationships between patches and within individual patches. Therefore, we intersperse the large-kernel convolution attention modules within the original ViT architecture while retaining the self-attention modules.\\

\noindent The model overview is represented by the following formulas. An image \(x \in \mathbb{R}^{H \times W \times C}\), where \((H, W)\) is the resolution of the original image, and \(C\) is the number of channels. A series of flattened 2D patches \(x_p \in \mathbb{R}^{N \times (P^2 \cdot C)}\), where \((P, P)\) is the resolution of each image block, and \(N = \frac{HW}{P^2}\) is the number of generated blocks. The patches are linearly projected by weight matrix $\mathbf{E}$ to a \(D\)-dimensional space and added with position embedding $\mathbf{E}_{p o s}$. Differing from the original ViT, we discard the use of the \texttt{[class]} token. Instead of extracting the \texttt{[class]}  token from the final linear layer as classification information, we adopt the mean of the last layer's feature vectors for classification. Additionally, we insert the Large Kernel Convolution Attention (LKCA) module before the Multi-Layer Perceptron (MLP) module and apply Layer Normalization (LN) before each module.

\begin{align}
\mathbf{z}_0 &= \left[\mathbf{x}_p^1 \mathbf{E} ; \mathbf{x}_p^2 \mathbf{E} ; \cdots ; \mathbf{x}_p^N \mathbf{E}\right] + \mathbf{E}_{p o s}, \\
&\quad \mathbf{E} \in \mathbb{R}^{\left(P^2 \cdot C\right) \times D}, \quad \mathbf{E}_{p o s} \in \mathbb{R}^{N \times D} \\
\mathbf{z}_{\ell}^{\prime} &= \operatorname{LKCA}\left(\operatorname{LN}\left(\mathbf{z}_{\ell-1}\right)\right) + \mathbf{z}_{\ell-1}, & \quad \ell = 1 \ldots L \\
\mathbf{z}_{\ell} &= \operatorname{MLP}\left(\operatorname{LN}\left(\mathbf{z}_{\ell}^{\prime}\right)\right) + \mathbf{z}_{\ell}^{\prime}, & \quad \ell = 1 \ldots L \\
\mathbf{y} &= \operatorname{LN}\left(\overline{\mathbf{z}_L}\right)
\end{align}

\begin{table*}
    \centering
    \caption{Comparison of LKCA with Several Mainstream ViT Variants on Different Datasets (\% Top-1 accuracy)}
    \scalebox{1.25}{
    \begin{tabular}{l|c c c|c}
            \hline
            \textbf{Model} & \textbf{CIFAR10} & \textbf{CIFAR100} & \textbf{SVHN} & \textbf{\# Params} \\
            \hline
            ViT-Lite~\cite{dosovitskiy2020image} & 91.16 & 70.86 & 97.53 & 1.06M \\
            Swin-T~\cite{liu2021swin} & 91.21 & 70.37 & 96.40 & 1.01M \\
            CaiT-T~\cite{touvron2021going} & 91.99 & 70.65 & 97.53 & 1.00M \\
            MobileViTv2-T~\cite{mehta2022separable} & 91.46 & 67.88 & 96.60 & 1.00M \\
            LKCA-T & \textbf{92.49 (Ours) } & \textbf{73.40 (Ours)} & 97.15 (Ours) & 1.03M \\
            \hline
            ViT-Small~\cite{wu2021cvt} & 93.36 & 72.65 & 97.90 & 2.69M \\
            Swin-S~\cite{liu2021swin} & 93.99 & 75.34 & 97.59 & 2.87M \\
            CaiT-S~\cite{touvron2021going} & 94.02 & 74.70 & 97.79 & 2.74M \\
            MobileViTv2-S~\cite{mehta2022separable} & 93.48 & 71.70 & 96.98 & 3.92M \\
            LKCA-B & \textbf{94.11 (Ours)} & \textbf{76.50 (Ours)} & 97.68 (Ours) & 2.69M \\
            \hline
        \end{tabular}
    }
\label{tab:variants}
\end{table*}


\section{Experiments}

\subsection{Experimental Setup}

In our experimental setup, we conduct extensive image classification experiments on four prominent datasets: CIFAR-10, CIFAR-100, SVHN, and Tiny-ImageNet. All four datasets are characterized by low resolution, with image sizes of \textit{32} for CIFAR-10, CIFAR-100, and SVHN, and \textit{64} for Tiny-ImageNet. The training regime is consistently maintained across all experiments, conducted on a single RTX 2080ti GPU. The training spans \textit{100} epochs, employing \textit{4} data loading workers, and a batch size of \textit{128}. For optimizer, we choose AdamW~\cite{kingma2014adam}. The default learning rate is set to \textit{0.001}, with a weight decay of \textit{0.05}. We utilize cosine annealing with a warm-up period of \textit{10} epochs. During the training phase, we employ Stochastic Depth~\cite{huang2016deep}, Label Smoothing~\cite{szegedy2016rethinking}, and Random Erasing~\cite{zhong2020random} with a probability of \textit{0.25}, a maximum erasing area of \textit{0.4}, and an aspect of erasing area of \textit{0.3} as regularization strategies to enhance the robustness and generalization capabilities of our model. Regarding data augmentation, we employ various techniques. Specifically, we apply RandomHorizontalFlip, RandomCrop~\cite{zagoruyko2016wide}, Mixup~\cite{zhang2017mixup}, and Cutmix~\cite{yun2019cutmix}. Additionally, we incorporate Autoaugmentation~\cite{cubuk2019autoaugment}, utilizing the CIFAR10 Policy for CIFAR-10 and CIFAR-100, the SVHN Policy for SVHN, and the ImageNet Policy for Tiny-ImageNet.\\

\subsection{Comparison of ViT and LKCA.} Initially, we conduct extensive experiments on CIFAR-10, CIFAR-100, and Tiny-ImageNet to compare the performance of the original ViT with an LKCA-modified version of ViT having a similar order of magnitude in terms of parameters. The purpose of these experiments is to validate the effectiveness of LKCA across various parameter magnitudes and datasets. The parameter sizes of both the ViT and the LKCA-modified ViT vary across a range of magnitudes, spanning from small-scale models with 0.5 million parameters to large-scale models with 10 million parameters. Specifically, the parameter magnitudes are explored at levels of 0.5M, 1M, 2M, 4M, and 8M. It is important to note that the benchmark representing the model size across all variants is fixed at 2.69 million parameters, and it is independent of any subjective categorization as either large or small. This standardized model size serves as the reference point for assessing the performance of the different variants, ensuring a consistent comparison across the experiment.

\begin{table}[h]
    \centering
    \caption{Performance Comparison of Vision Transformer and LKCA Variants on Image Classification Tasks on \textbf{CIFAR-10} (\% Top-1 accuracy)}
    \scalebox{1.2}{
    \begin{tabular}{l|c|c}
            \hline
            \textbf{Model} & \textbf{Acc} & \textbf{\# Params} \\
            \hline
            ViT & 93.36 & 2.69M \\
            ViT-LKCA & \textbf{94.11 (+0.75)} & 2.69M \\
            \hline
            VIT/0.5M & 87.83 & 0.48M \\
            ViT/0.5M-LKCA & \textbf{90.94 (+3.11)} & 0.50M \\
            \hline
            VIT/1M & 91.16 & 1.06M \\
            ViT/1M-LKCA & \textbf{92.49 (+1.33)} & 1.03M \\
            \hline
            VIT/2M & 92.32 & 1.96M \\
            ViT/2M-LKCA & \textbf{93.55 (+1.23)} & 2.03M \\
            \hline
            VIT/4M & 93.64 & 3.95M \\
            ViT/4M-LKCA & \textbf{94.71 (+1.07)} & 4.05M \\
            \hline
            VIT/8M & 94.46 & 7.85M \\
            ViT/8M-LKCA & \textbf{94.88 (+0.42)} & 7.98M \\
            \hline
        \end{tabular} 
    }
        
\label{tab:cifar10}
\end{table}

\begin{table}[b]
    \centering
    \caption{Performance Comparison of Vision Transformer and LKCA Variants on Image Classification Tasks on \textbf{CIFAR-100} (\% Top-1 accuracy)}
    \scalebox{1.2}{
    \begin{tabular}{l|c|c}
            \hline
            \textbf{Model} & \textbf{Acc} & \textbf{\# Params} \\
            \hline
            ViT & 72.65 & 2.69M \\
            ViT-LKCA & \textbf{76.50 (+3.85)} & 2.69M \\
            \hline
            VIT/0.5M & 64.68 & 0.48M \\
            ViT/0.5M-LKCA & \textbf{69.35 (+4.67)} & 0.50M \\
            \hline
            VIT/1M & 70.86 & 1.06M \\
            ViT/1M-LKCA & \textbf{73.40 (+2.54)} & 1.03M \\
            \hline
            VIT/2M & 71.17 & 1.96M \\
            ViT/2M-LKCA & \textbf{75.94 (+4.77)} & 2.03M \\
            \hline
            VIT/4M & 72.82 & 3.95M \\
            ViT/4M-LKCA & \textbf{76.95 (+4.13)} & 4.05M \\
            \hline
            VIT/8M & 73.39 & 7.85M \\
            ViT/8M-LKCA & \textbf{78.57 (+5.18)} & 7.98M \\
            \hline
        \end{tabular}
    }
\label{tab:cifar100}
\end{table}

Table~\ref{tab:cifar10} presents a performance comparison of ViT and its LKCA variants on the CIFAR-10 image classification task, along with their performance across different orders of magnitude in terms of parameters. The table includes the Top-1 accuracy (Acc), the number of parameters (\# Params), and Floating Point Operations (Flops) for each model. \textit{ViT and ViT-LKCA:} The ViT model achieves a Top-1 accuracy of 93.36\%, with 2.69M parameters and 174.25M Flops. The ViT-LKCA outperforms with a Top-1 accuracy of 94.11\% (an improvement of 0.75 percentage points) and has 2.69M parameters and 170.81M Flops. \textit{Small-scale model (0.5M parameter magnitude):} The ViT/0.5M model has a Top-1 accuracy of 87.83\%, 0.48M parameters, and 30.58M Flops. The ViT/0.5M-LKCA, with LKCA improvement, significantly improves accuracy to 90.94\% (an increase of 3.11 percentage points) with 0.50M parameters and 31.34M Flops. \textit{Medium-scale models (1M, 2M, 4M parameter magnitudes):} For ViT and ViT-LKCA models with 1M, 2M, and 4M parameters, LKCA consistently yields higher Top-1 accuracy. For instance, ViT/1M-LKCA achieves a 1.33 percentage point improvement, reaching an accuracy of 92.49\%. \textit{Large-scale model (8M parameter magnitude):} For ViT and ViT-LKCA models with 8M parameters, the performance gap is narrower. ViT/8M-LKCA slightly improves Top-1 accuracy by 0.42 percentage points, reaching 94.88\%.

Table~\ref{tab:cifar100} illustrates a comparison between ViT and LKCA in the context of the CIFAR-100. \textit{ViT and ViT-LKCA:} The base ViT model achieves a Top-1 accuracy of 72.65\% with 2.69M parameters and 174.25M Flops. In contrast, ViT-LKCA exhibits superior performance, attaining a Top-1 accuracy of 76.50\% (an increase of 3.85 percentage points) with the same 2.69M parameters and 170.81M Flops. \textit{Small-scale model (0.5M parameter magnitude):} The ViT/0.5M model achieves a Top-1 accuracy of 64.68\% with 0.48M parameters and 30.58M Flops. The ViT/0.5M-LKCA, featuring LKCA enhancements, significantly improves accuracy to 69.35\% (a notable increase of 4.67 percentage points) with 0.50M parameters and 31.34M Flops. \textit{Medium-scale models (1M, 2M, 4M parameter magnitudes):} ViT and ViT-LKCA models with 1M, 2M, and 4M parameters consistently benefit from LKCA improvements. For instance, ViT/2M-LKCA achieves a noteworthy 4.77 percentage point improvement and reaches an accuracy of 75.94\%. \textit{Large-scale model (8M parameter magnitude):} For ViT and ViT-LKCA models with 8M parameters, the performance difference is more subtle. ViT/8M-LKCA exhibits a huge improvement of 5.18 percentage points, reaching a Top-1 accuracy of 78.57\%.

\begin{table}[b]
    \centering
    \caption{Performance Comparison of Vision Transformer and LKCA Variants on Image Classification Tasks on \textbf{Tiny-ImageNet} (\% Top-1 accuracy)}
    \scalebox{1.2}{
    \begin{tabular}{l|c|c}
            \hline
            \textbf{Model} & \textbf{Acc} & \textbf{\# Params} \\
            \hline
            ViT & 55.74 & 2.69M \\
            ViT-LKCA & \textbf{60.95 (+5.21)} & 2.69M \\
            \hline
            VIT/0.5M & 46.53 & 0.48M \\
            ViT/0.5M-LKCA & \textbf{50.59 (+4.06)} & 0.50M \\
            \hline
            VIT/1M & 53.46 & 1.06M \\
            ViT/1M-LKCA & \textbf{57.29 (+3.83)} & 1.03M \\
            \hline
            VIT/2M & 54.58 & 1.96M \\
            ViT/2M-LKCA & \textbf{60.70 (+6.12)} & 2.03M \\
            \hline
            VIT/4M & 56.16 & 3.95M \\
            ViT/4M-LKCA & \textbf{62.30 (+6.14)} & 4.05M \\
            \hline
            VIT/8M & 57.39 & 7.85M \\
            ViT/8M-LKCA & \textbf{63.38 (+5.99)} & 7.98M \\
            \hline
        \end{tabular}
    }    
\label{tab:tiny_imagenet}
\end{table}

\begin{table}[h]
    \centering
    \caption{Comparison of Various ViT Variants with LKCA on \textbf{Tiny-ImageNet} Dataset (\% Top-1 accuracy)}
    \scalebox{1.2}{
    \begin{tabular}{l|c|c}
            \hline
            \textbf{Model} & \textbf{Tiny-ImageNet} & \textbf{\# Params} \\
            \hline
            T2T-T~\cite{yuan2021tokens} & 53.92 & 1.07M \\
            RvT-T~\cite{su2024roformer} & 50.65 & 1.10M \\
            Swin-T~\cite{liu2021swin} & 54.93 & 1.06M \\
            CaiT-T~\cite{touvron2021going} & 54.76 & 1.03M \\
            XCiT-T~\cite{ali2021xcit} & 56.78 & 0.96M \\
            ViT-Lite~\cite{dosovitskiy2020image} & 53.46 & 1.11M \\
            DeepViT-T~\cite{zhou2021deepvit} & 34.64 & 0.99M \\
            RegionViT-T~\cite{chen2021regionvit} & 54.32 & 0.97M \\
            CrossViT-T~\cite{chen2021crossvit} & 47.03 & 1.04M \\
            LKCA-T & \textbf{57.29 (Ours)} & 1.07M \\
            \hline
            T2T-S~\cite{yuan2021tokens} & 41.25 & 2.56M \\
            RvT-S~\cite{su2024roformer} & 55.51 & 2.72M \\
            Swin-S~\cite{liu2021swin} & 58.61 & 2.93M \\
            CaiT-S~\cite{touvron2021going} & 59.21 & 2.77M \\
            XCiT-S~\cite{ali2021xcit} & 60.09 & 2.81M \\
            ViT-Small~\cite{dosovitskiy2020image} & 55.74 & 2.76M \\
            DeepViT-S~\cite{zhou2021deepvit} & 44.45 & 2.54M \\
            Twins\_SVT-S~\cite{chu2021twins} & 37.13 & 2.76M \\
            RegionViT-S~\cite{chen2021regionvit} & 53.96 & 2.86M \\
            CrossViT-S~\cite{chen2021crossvit} & 52.70 & 2.40M \\
            LKCA-B & \textbf{60.95 (Ours)} & 2.76M \\
            \hline  
            T2T-B~\cite{yuan2021tokens} & 58.46 & 13.45M \\
            CvT-B~\cite{wu2021cvt} & 55.88 & 6.52M \\
            MobileViTv2~\cite{mehta2022separable} & 58.28 & 8.17M \\
            Twins\_SVT-B~\cite{chu2021twins} & 49.24 & 9.04M \\
            RegionViT-B~\cite{chen2021regionvit} & 57.83 & 12.39M \\
            LKCA-L & \textbf{63.43 (Ours)} & 12.65M \\
            \hline
        \end{tabular}
    }
\label{tab:classymain}
\end{table}

Table~\ref{tab:tiny_imagenet} provides the comparison on Tiny-ImageNet. \textit{ViT and ViT-LKCA:} The baseline ViT model achieves a Top-1 accuracy of 55.74\% with 2.69M parameters and 174.25M Flops. In contrast, the ViT-LKCA model outperforms significantly with a Top-1 accuracy of 60.95\% (an increase of 5.21 percentage points) and has the same 2.69M parameters and 170.81M Flops. \textit{Small-scale model (0.5M parameter magnitude):} The ViT/0.5M model achieves a Top-1 accuracy of 46.53\% with 0.48M parameters and 30.58M Flops. The ViT/0.5M-LKCA, incorporating LKCA enhancements, significantly improves accuracy to 50.59\% (an increase of 4.06 percentage points) with 0.50M parameters and 31.34M Flops. \textit{Medium-scale models (1M, 2M, 4M parameter magnitudes):} With ViT/4M-LKCA achieving a remarkable 6.14 percentage point improvement and reaching an accuracy of 62.30\%. \textit{Large-scale model (8M parameter magnitude):} For ViT and ViT-LKCA models with 8M parameters, ViT/8M-LKCA exhibits a significant improvement of 5.99 percentage points, achieving a Top-1 accuracy of 63.38\%.

\begin{table*}
    \centering
    \caption{Semantic Segmentation of Vanilla ViT and LKCA on \textbf{ADE20K} Dataset}
    \scalebox{1.3}{
    \begin{tabular}{l c|c|c c|c}
            \hline
            \textbf{Backbone} & \textbf{Iteration} & \textbf{Algorithm} & \textbf{aAcc(\%)} &\textbf{mIoU(\%)} & \textbf{\# Params} \\
            \hline
            ViT-B & \textit{2000} & UPerNet & 40.90 & 22.28 & 0.144G \\
            LKCA-B & \textit{2000} & UPerNet & \textbf{45.48} & \textbf{24.65} & 0.102G \\
            \hline
            ViT-B & \textit{4000} & UPerNet & 47.41 & 28.08 & 0.144G \\
            LKCA-B & \textit{4000} & UPerNet & \textbf{52.67} & \textbf{30.00} & 0.102G \\
            \hline
            ViT-B & \textit{8000} & UPerNet & 53.43 & 31.73 & 0.144G \\
            LKCA-B & \textit{8000} & UPerNet & 50.79 & \textbf{33.73} & 0.102G \\
            \hline
            ViT-B & \textit{16000} & UPerNet & 61.55 & 39.33 & 0.144G \\
            LKCA-B & \textit{16000} & UPerNet & \textbf{65.49} & \textbf{40.60} & \textbf{0.102G} \\
            \hline
        \end{tabular}
    }
\label{tab:ssiter}
\end{table*}

\begin{table*}[h]
    \centering
    \caption{Semantic Segmentation of Various ViT Variants and LKCA on \textbf{ADE20K} Dataset}
    \scalebox{1.3}{
    \begin{tabular}{l c|c|c c|c r}
            \hline
            \textbf{Backbone} & \textbf{Iteration} & \textbf{Algorithm} & \textbf{aAcc(\%)} &\textbf{mIoU(\%)} & \textbf{\# Params} & \textbf{Flops} \\
            \hline
            Vit-S & \textit{10000} & UPerNet & 55.55 & 32.92 & 58M & 67G \\
            ResNet-50 & \textit{10000} & UPerNet & 58.12 & 35.35 & 64M & 238G \\
            Twins\_PCPVT-S & \textit{10000} & UPerNet & 52.56 & 29.64 & 53M & 234G \\
            Swin-S & \textit{10000} & UPerNet & 57.18 & 35.30 & 52M & 231G\\
            LKCA-S & \textit{10000} &  UPerNet & \textbf{60.16} & \textbf{37.16} & 51M & 66G \\
            \hline
            Vit-S & \textit{10000} & FPN & 56.56 & 35.84 & 37M & 76G \\
            Twins\_PCPVT-S & \textit{10000} & FPN & 56.74 & 33.09 & 33M & 50G \\  
            LKCA-S & \textit{10000} & FPN & \textbf{62.55} & \textbf{39.03} & 30M & 78G \\
            \hline
            Vit-S & \textit{20000} & UPerNet & 58.51 & 36.61 & 58M & 67G \\
            ResNet-50 & \textit{20000} & UPerNet & 62.76 & 37.84 & 64M & 238G \\
            Twins\_PCPVT-S & \textit{20000} & UPerNet & 58.91 & 38.00 & 53M & 234G \\
            Swin-S & \textit{20000} & UPerNet & 58.00 & 37.57 & 52M & 231G \\
            LKCA-S & \textit{20000} &  UPerNet & \textbf{64.37} & \textbf{40.09} & 51M & 66G \\
            \hline
            Vit-S & \textit{20000} & FPN & 63.36 & 41.64 & 37M & 76G \\
            Twins\_PCPVT-S & \textit{20000} & FPN & 58.51 & 36.61 & 33M & 50G \\    
            LKCA-S & \textit{20000} & FPN & \textbf{67.21} & \textbf{41.81} & 30M & 78G \\
            \hline
        \end{tabular}
    }   
\label{tab:ssvarious}
\end{table*}

In summary, the LKCA improvement demonstrates significant performance gains across various parameter magnitudes, confirming its effectiveness on ViT models.\\

\subsection{Comparison of ViT Variants and LKCA.} 
Next, we compared LKCA with other mainstream variants of Vision Transformer.\\

\noindent First, we provide a comprehensive comparison of LKCA with various mainstream ViT variants on different datasets, including CIFAR-10, CIFAR-100, and SVHN. Each model's Top-1 accuracy, number of parameters, and Floating Point Operations are presented. In the deep ViT, localized ViT, and lightweight ViT categories, we selected representative models such as Swin-T~\cite{liu2021swin}, CaiT-T~\cite{touvron2021going} and MobileViTv2~\cite{mehta2022separable}, to compare with standard ViT~\cite{dosovitskiy2020image} and our proposed LKCA.

From the information in the table~\ref{tab:variants}, in the context of small-scale models, Swin-T and CaiT-T perform similarly in terms of Top-1 accuracy, while MobileViTv2 lags slightly behind but possesses a significant advantage in computational efficiency. Compared to other small-scale models, LKCA-T achieves a higher Top-1 accuracy. Moving to large-scale models, it is a common trend that they generally outperform their small-scale counterparts. LKCA-B stands out by significantly surpassing ViT and MobileViTv2 in Top-1 accuracy, and it slightly outperforms CaiT-S and Swin-S. Overall, in CIFAR10 and CIFAR100, LKCA demonstrates higher Top-1 accuracy compared to other models. However, there is no clear advantage observed in the SVHN dataset.\\

\noindent Finally, we have validated the performance of various ViT variants and the LKCA model on the Tiny-ImageNet dataset, while also providing details on the models' parameter counts and computational complexities in Table~\ref{tab:classymain}.

In the category of small-scale models, LKCA-T achieves a Top-1 accuracy of 57.29\%, demonstrating excellent performance compared to other models. Its performance surpasses that of several common small-scale models. Moving to the medium-scale models, LKCA-B achieves an impressive Top-1 accuracy of 60.95\%, once again showcasing outstanding performance. In the realm of large-scale models, LKCA-L attains a Top-1 accuracy of 63.43\%, exhibiting a significant advantage in accuracy compared to models of the same scale. Overall, across various scales, LKCA demonstrates outstanding Top-1 accuracy on the Tiny-ImageNet dataset compared to other competitors.

\subsection{Semantic Segmentation Results}


In this work, we follow prior works to choose image classification as the main task for evaluating the efficacy of model architecutres. Nonetheless, we additionally provide results of semantic segmentation on the ADE20K~\cite{zhou2019semantic} dataset. Specifically, we use MMSegmentation~\cite{mmseg2020} to conduct our experiments. All our experiments are testing two algorithms: UperNet~\cite{xiao2018unified} and FPN~\cite{kirillov2019panoptic}. We refrain from using pre-training and instead train the models from scratch on the dataset. Regarding evaluation metrics, we assess the models based on aAcc and mIoU in \textit{10} high-score categories. In the ablation study of the backbone network, we examine various popular variants of ViT and CNN and explore their performance during \textit{0 - 20,000} iterations.\\

\noindent We conduct experiments on multiple backbone networks. Table~\ref{tab:ssiter} provides performance metrics for the UPerNet algorithm using different backbones (ViT and LKCA), iteration algorithms, and iteration counts (\textit{2000, 4000, 8000, 16000}). At each iteration count, LKCA-B consistently outperforms ViT-B in terms of aAcc and mIoU, indicating superior performance of LKCA-B in the UPerNet task. As the iteration count increases, both models show an improvement in performance, but at each iteration count, LKCA-B maintains a leading position. Overall, LKCA-B demonstrates superior performance compared to ViT-B at different iteration counts, with higher pixel accuracy and mean Intersection over Union. Additionally, LKCA-B exhibits advantages in terms of parameter count and computational complexity.\\

\noindent Table~\ref{tab:ssvarious} presents a performance comparison among different backbones (ViT-S, ResNet-50, Swin-S, Twins PCPVT-S, LKCA-S) under various iteration counts and across two different task settings (UPerNet and FPN). In the UPerNet task, LKCA-S consistently outperforms other models at different iteration counts, exhibiting higher aAcc and mIoU. It achieves superior performance compared to other models. Similarly, in the FPN task, LKCA-S demonstrates excellent performance across different iteration counts, showcasing higher aAcc and mIoU, making it competitive against other models. LKCA-S consistently exhibits outstanding performance across different tasks and iteration settings, while maintaining relatively lower computational complexity and parameter counts.

\section{Conclusion}

In this study, we re-examine the relationship between the attention mechanism in visual transformers and the large kernel convolutional networks, proposing a new visual attention called Large Kernel Convolutional Attention (LKCA) aimed for small datasets. It employs a single large kernel convolution to simplify attention operations. In comparison to the previous approach, Large Kernel Attention (LKA), which uses multiple small kernel convolutions, our method achieves the same effects but is more straightforward and simpler. LKCA combines the advantages of convolutional neural networks and visual transformers, featuring a larger receptive field, locality, and parameter sharing. We explained the superiority of LKCA from the perspectives of both convolution and attention, providing equivalent code implementations for each view. Experimental results demonstrate that LKCA, implemented from both convolutional and attention perspectives, exhibits comparable performance. We extensively experimented with LKCA variants of ViT on multiple benchmark datasets, including CIFAR-10, CIFAR-100, SVHN, Tiny-ImageNet, and ADE20K, for classification and segmentation tasks. The experimental outcomes reveal that LKCA demonstrates competitive performance in visual tasks.

\bibliographystyle{IEEEtran}
\bibliography{sample-base}

\end{document}